\def\cvprPaperID{0000}
\def\confName{CVPR}
\def\confYear{2023}

\def\paperTitle{How to Efficiently Adapt Large Segmentation Model(SAM) to Medical Images}

\def\authorBlock{
    Xinrong Hu$^{1}$ \qquad
    Xiaowei Xu$^{2}$ \qquad
    Yiyu Shi$^{1}$ \\
    $^{1}$ Department of Computer Science and Engineering, University of Notre Dame \\
     $^{2}$ Guangdong Cardiovascular Institute, Guangdong
Provincial People’s Hospital \\
    {\tt\small \{xhu7, xxu8, yshi4\}@nd.edu}
}

\newif\ifreview 
\newif\ifarxiv \newcommand{\arxiv}{\arxivtrue}
\newif\ifcamera 
\newif\ifrebuttal 

\arxiv 

\pdfoutput=1
\documentclass[10pt,twocolumn,letterpaper]{article}
\ifreview \usepackage[review]{cvpr} \fi
\ifarxiv \usepackage[pagenumbers]{cvpr} \fi
\ifrebuttal \usepackage[rebuttal]{cvpr} \fi
\ifcamera \usepackage{cvpr} \fi

\usepackage{graphicx}
\usepackage{amsmath}
\usepackage{amssymb}
\usepackage{booktabs}

\usepackage{times}
\usepackage{microtype}
\usepackage{epsfig}
\usepackage[table,xcdraw]{xcolor}
\usepackage{caption}
\usepackage{float}
\usepackage{placeins}
\usepackage{color, colortbl}
\usepackage{stfloats}
\usepackage{enumitem}
\usepackage{tabularx}
\usepackage{xstring}
\usepackage{multirow}
\usepackage{xspace}
\usepackage{url}
\usepackage{subcaption}
\usepackage{xcolor}
\usepackage[hang,flushmargin]{footmisc}

\ifcamera \usepackage[accsupp]{axessibility} \fi

% ...

\ifarxiv  \fi

\newcommand{\R}[1]{{%
    \textbf{%
        \ifstrequal{#1}{1}{\textcolor{red}{R#1}}{%
        \ifstrequal{#1}{2}{\textcolor{blue}{R#1}}{%
        \ifstrequal{#1}{3}{\textcolor{magenta}{R#1}}{%
        \ifstrequal{#1}{4}{\textcolor{teal}{R#1}}{%
                           \textcolor{cyan}{R#1}%
        }}}}%
    }%
}}

\usepackage{xr-hyper}

\makeatletter
\newcommand*{\addFileDependency}[1]{
  \typeout{(#1)}
  \@addtofilelist{#1}
  \IfFileExists{#1}{}{\typeout{No file #1.}}
}

\makeatother

\usepackage[pagebackref,breaklinks,colorlinks]{hyperref}
\usepackage[capitalize]{cleveref}
\crefname{section}{Sec.}{Secs.}
\crefname{table}{Table}{Tables}
\crefname{figure}{Fig.}{Figs.}

\begin{document}
\title{\paperTitle}
\author{\authorBlock}
\maketitle
\begin{abstract}
The emerging scale segmentation model, Segment Anything (SAM), exhibits impressive capabilities in zero-shot segmentation for natural images.
However, when applied to medical images, SAM suffers from noticeable performance drop.
To make SAM a real ``foundation model" for the computer vision community, it is critical to find an efficient way to customize SAM for medical image dataset.
In this work, we propose to freeze SAM encoder and finetune a lightweight task-specific prediction head, as most of weights in SAM are contributed by the encoder.
In addition, SAM is a promptable model, while prompt is not necessarily available in all application cases, and precise prompts for multiple class segmentation are also time-consuming. 
Therefore, we explore three types of prompt-free prediction heads in this work, include ViT, CNN, and linear layers. 
For ViT head, we remove the prompt tokens in the mask decoder of SAM, which is named AutoSAM.
AutoSAM can also generate masks for different classes with one single inference after modification.
To evaluate the label-efficiency of our finetuning method, we compare the results of these three prediction heads on a public medical image segmentation dataset with limited labeled data.
Experiments demonstrate that finetuning SAM significantly improves its performance on medical image dataset, even with just one labeled volume. 
Moreover, AutoSAM and CNN prediction head also has better segmentation accuracy than training from scratch and self-supervised learning approaches when there is a shortage of annotations.
The codes are available at \href{https://github.com/xhu248/AutoSAM}{https://github.com/xhu248/AutoSAM}.
\end{abstract}

\begin{figure}
    \centering
    \includegraphics[width=0.9\linewidth]{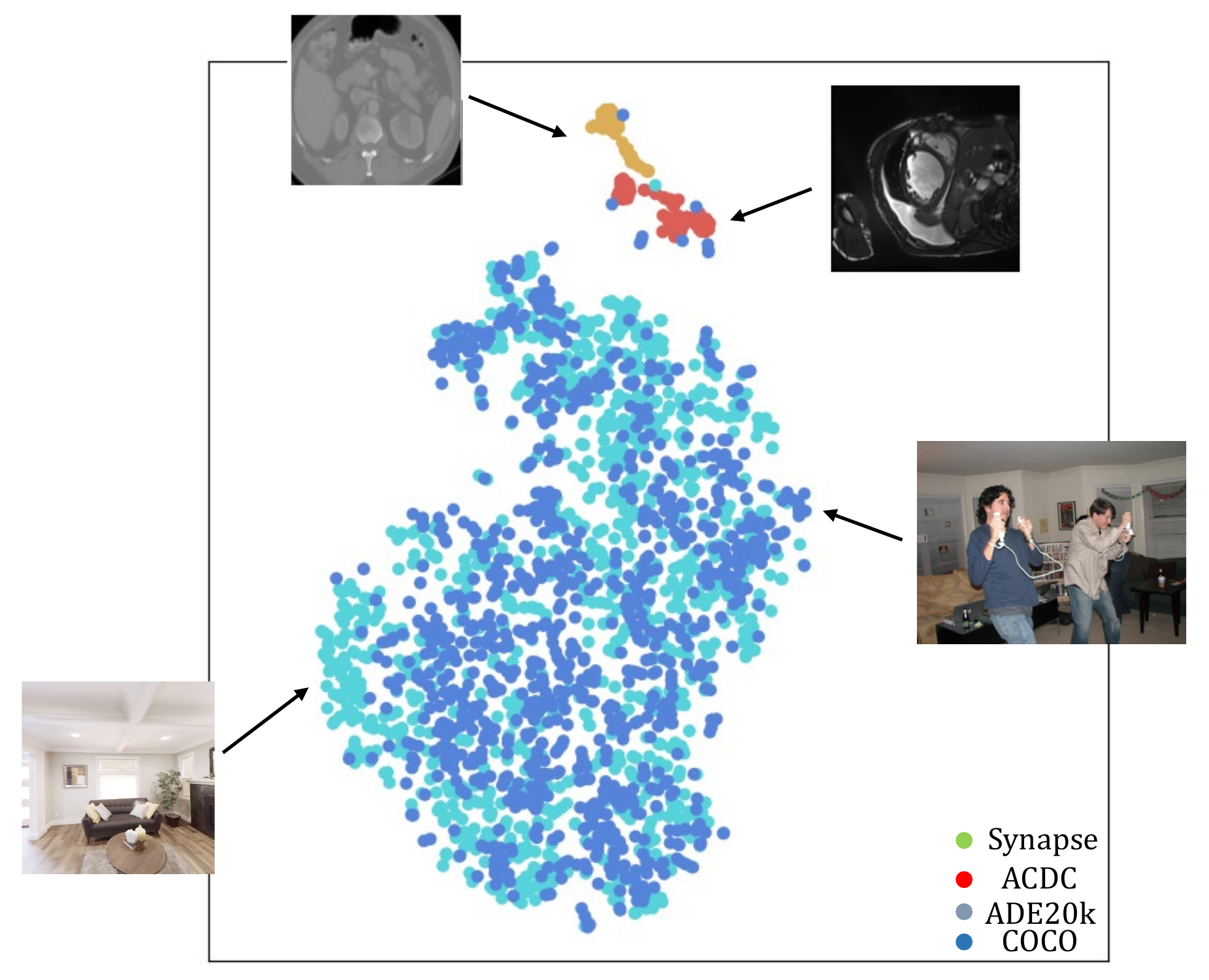}
    \caption{T-SNE plot of embeddings encoded by SAM's image encoder from four datasets. The four datasets are Synapse\cite{multi-atlas}, ACDC\cite{bernard2018deep}, ADE20K\cite{zhou2017scene}, and COCO\cite{lin2014microsoft}. As is showed, there is a apparent domain shift from natural images to medical images in latent space. This may explain why SAM fails to have good performance on unseen medical image datasets.}
    \label{fig1}
\end{figure}

\section{Introduction}
\label{sec:introduction}
The success of Generative Pre-trained Transformer(GPT)\cite{ouyang2022training,brown2020language,solaiman2019release} series models demonstrates that if trained on large scaled data, the performance of large language model on zero-shot and few-shot tasks in unseen domain is comparable with state of the arts.
Inspired by GPT, Segment Anything (SAM)\cite{kirillov2023segment} introduces a "foundation model" for image segmentation task. 
They collect 11 million images and design a semi-automatic data engine to yield on average $\sim$100 masks per image, thus 1 billion masks in total.
Then SAM trains a large promptable model with Vision Transformer\cite{dosovitskiy2020image} (ViT) backbone on this SAM-1B dataset. 
After being evaluated with various zero-shot tasks on over 23 datasets, SAM demonstrates promising generalization to most natural images.

However, as SAM draws attention in medical image domains, it is observed that SAM does not generalize well to medical images in zero-shot settings\cite{ma2023segment, he2023accuracy}. 
The challenges of transferring model trained with natural images to medical images can be attributed to two main factors: 1) \textbf{Large difference in appearance}: Natural images and medical images exhibit significant differences in terms of color, brightness, and contrast. Medical images often have distinct characteristics due to the imaging modalities used, such as CT scans, MRI, or ultrasound; 2) \textbf{Blurred boundaries of target objects}: Medical images frequently present blurred boundaries between different tissues and organs. Trained medical experts possess the necessary understanding of anatomical structures and can identify subtle boundaries that may not be apparent to models trained solely on natural images.
Considering the difficulty of collecting a medical segmentation dataset with comparable size as SAM-1B, it is critical to explore if there is knowledge in the pre-trained SAM that can be exploited for medical image segmentation.

Furthermore, prompt-based segmentation might not be well-suited for real-world application scenarios due to the following reasons: 1) Providing prompt for multi-class is time-consuming. 
For most public medical image segmentation challenges, it always require segmenting multiple classes simultaneously. Inputting accurate prompts for each class can become cumbersome especially when organs and tissues are small and adjacent to each other; 2) The segmentation performance is heavily dependent on the prompt quality. Crafting precise prompting needs expert domain-specific knowledge, which is not available for all circumstances.

With these limitations in mind, this paper proposes a straightforward way to finetune the SAM on medical image datasets, that is, freezing the weights of SAM encoder and adding a prediction head on it for training. 
The reason for freezing the weights is that SAM is a large model and most of the weights are contributed by the encoder. 
Finetuning both the encoder and the decoder is not only less accessible to all developers due to high hardware requirement, but also results in worse segmentation performance according to experiment results.
On the other hand, to improve SAM's feasibility for clinical applications, we replace the mask decoder in SAM with a prediction head requiring no prompts for both training and inference.
Three different types of prediction heads are evaluated in this paper, including Vision Transformer (ViT), Convolutional Neural Network (CNN), and Linear layer.
The ViT prediction head is adapted from SAM mask decoder, which is named as AutoSAM, composed of lightweight cross-attention modules and transposed convolutional layers.
We remove the prompt tokens and replicate image embedding as well as other auxiliary embeddings so that the decoder can generate multiple masks for different classes at the same time.

In order to showcase the label-efficiency of our method, we conduct experiments in a few-shot learning setting, where the model is finetuned using only 1 or 5 labeled MRI scans. 
The results obtained on a publicly available medical image segmentation dataset highlight the significant improvement achieved through customizing pre-trained SAM compared with zero-shot prompt-drive SAM.
Moreover, our approach outperforms both training from scratch and state-of-the-art self-supervised learning methods by a substantial margin, highlighting the potential of SAM's application to medical domains.


\section{Related Works}
\label{sec:related}
\textbf{Large Vision Models}
After the emerging of large language models (LLM), some works are devoted to introducing image in LLM to accomplish multi-modality tasks. 
For example, CLIP\cite{radford2021learning} and ALIGN\cite{jia2021scaling} utilizes contrastive learning to align web images and their captions in embedding space. 
They find this simple pre-training task can be generalized well to other zero-shot downstream tasks, like object classification and action recognition in videos.
Also, DALL-E\cite{ramesh2021zero} achieves great generalization with a scale autoregressive transformer for zero-shot text-to-image generation.
However, these large scale vision model fails to tackle the wide range of all computer vision tasks, like image segmentation.
The difficulty of obtaining label mask is the key for a large image segmentation model.
SAM (Segment Anything)\cite{kirillov2023segment} is the first work to develop a promptable segmentation model and pre-train it on a broad  dataset by themselves. 
Given suitable prompts, SAM is capable of generating promising mask for target object without task specific training. 
On the other hand, DINOv2\cite{oquab2023dinov2} scales the pre-training of a ViT model in terms of data and model size, in order to produce all-purpose visual features, with which the finetuning of downstream tasks can be much easier.

\textbf{Customizing Large Models for Medical Images}
This family of works mainly focus on finetuning SAM for specific segmentation dataset, as SAM shows significant performance degradation on medical images.
MedSAM\cite{ma2023segment} simply finetune SAM decoder with prompt generated from label masks on over 30 medical image datasets, and results show improvement over zero-shot predictions generated with prompts.
Kaidong Zhang et al. \cite{zhang2023customized} applies the low-rank-based finetuning strategy to the SAN encoder and train it together with SAM decoder to customize SAM to abdominal segmentation tasks.
Junde Wu et al. \cite{wu2023medical} freezes weights of SAM model and adds trainable adaptation module in SAM to reduce the cost of re-training.
\begin{figure*}
    \centering
    \includegraphics[width=0.85\linewidth]{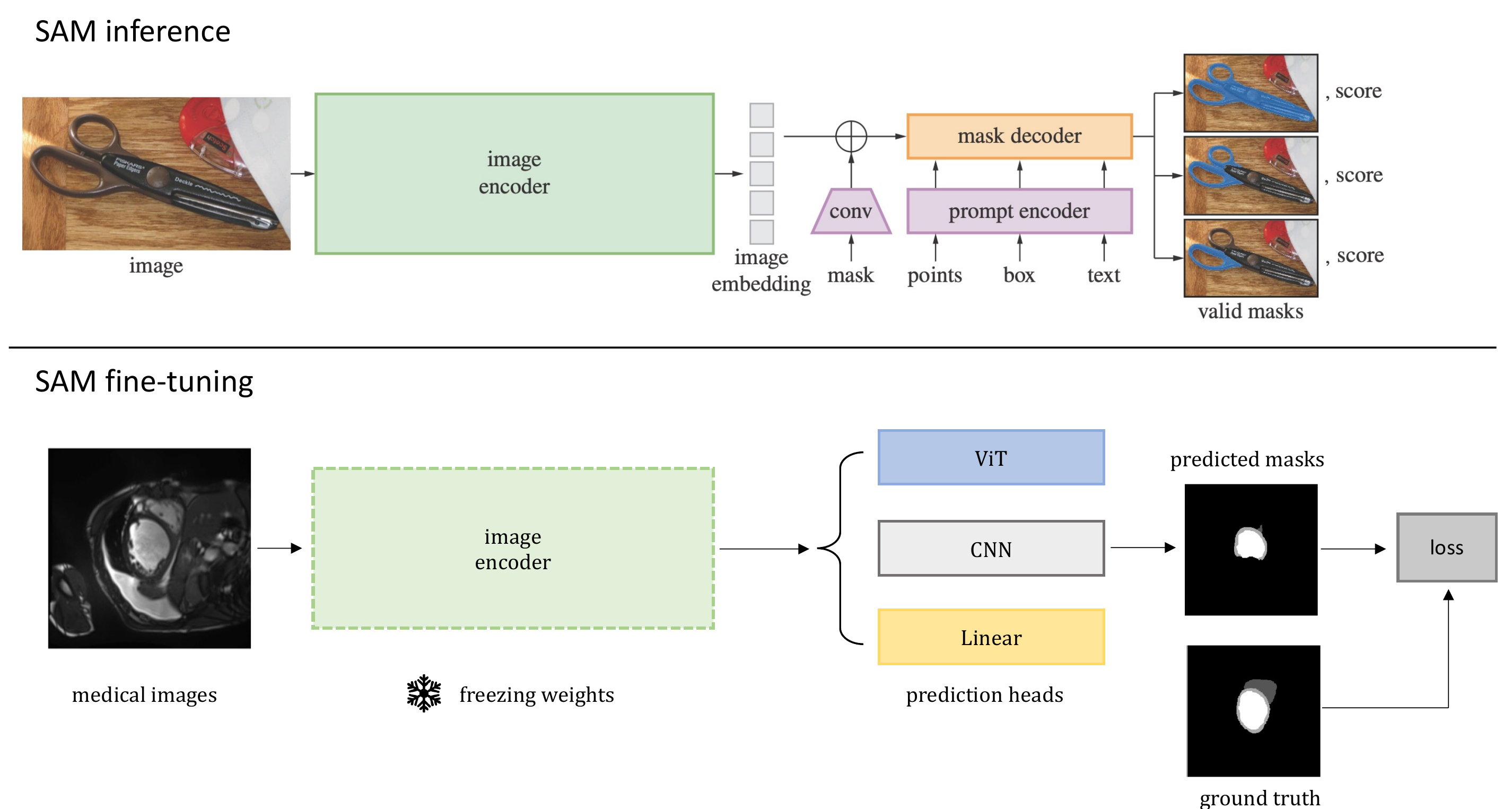}
    \caption{Comparisons of SAM inference process and our SAM finetuing process. We freeze the weights in SAM encoder, and adds various of prediction heads to generate segmentation mask without prompts, including Vision Transformer (ViT), CNN, and linear layer. Also, our model can generate masks of difference target objects.}
    \label{fig2}
\end{figure*}

\section{Methods}
\label{sec:methods}

\subsection{Background}

Firstly, we will give a brief introduction to the SAM model as background knowledge.
There are three major components in SAM, image encoder, prompt encoder, and mask decoder.
The image encoder has the same architecture as Vision Transformer (ViT) \cite{dosovitskiy2020image}, and is pre-trained with MAE\cite{he2022masked} on their own collected SAM-1B dataset. 
They provide weights of three different scale image encoder, ViT-h, ViT-l, and Vit-b, as options for trade-off between real-time performance and accuracy.
The image encoder takes any size of input images, and reshape it to 1024*1024. 
Then the images are converted to sequential patch embeddings with patch size 16*16 and embedding size 256.
After several transformer blocks with window attention and residual propagation, the output
of image encoder has dimension of (64x64, 256).
The prompt encoder support both sparse prompts (points, boxes, text) and dense prompts (masks).
Sparse prompts are projected into prompt tokens and concatenated with image embedding, while dense prompts are embedded using convolutions and summed element-wise with the image embedding.
The mask decoder firstly applies a two-way attention module on output token, prompt token, and image embedding.
Then the image embedding are upsampled by two transposed convolutional layer, and
the prediction is made on point-wise product between the upscaled image embedding and  output token.
More details of the mask decoder will be discussed in the following section.

\begin{figure*}
    \centering
    \includegraphics[width=0.9\linewidth]{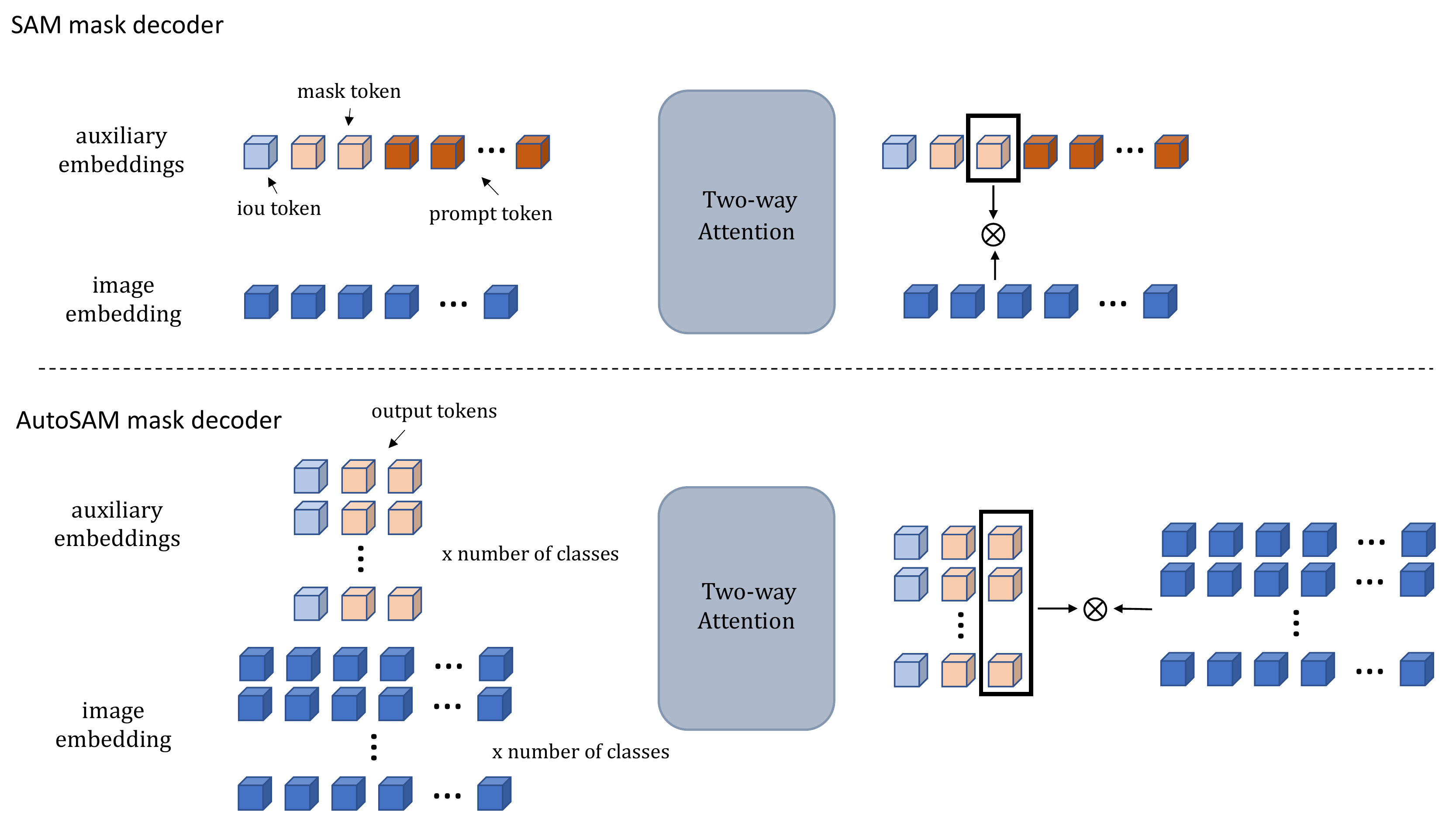}
    \caption{Illustration of mask decoder in SAM and AutoSAM. AutoSAM removes the prompt token so that it requires no input prompt, from which the name "Auto" come. To enable multi-class segmentation at the same time, AutoSAM copies the pair of auxiliary embeddings and image embedding by the number of classes. Parallel computing can reduce the computation overhead associated with the duplicated embedding. The two-way attention includes self-attention blocks and cross-attention blocks.  }
    \label{fig3}
\end{figure*}

\subsection{Prediction Head}
To adapt SAM to a certain medical image dataset in an efficient way, we keep the weights in SAM encoder and appends an additional task-specific prediction head for finetuning. 
Also, we design the prediction head to be not promptable and the only input is the image embedding from SAM encoder.
We explore three most common architecture types, ViT, CNN, and linear layer.

\textbf{Vision Transformer} We notice that the original mask decoder in SAM has ViT backbones, thus we can make a light modification on it so that the prediction head is not only non-promptable but is also able to utilize the weights in SAM mask decoder.
As is illustrated in Fig. \ref{fig2}, for SAM decoder, asides from prompt token and image embedding, there are also trainable output tokens, including mask token for generating mask and IoU token for predicting the confidence for the mask.
Furthermore, the mask tokens comprise foreground mask token and background mask token.
The output tokens are concatenated with prompt tokens, which we name as auxiliary embeddings.
In the two-way attention module, each layer performs both self-attention and cross-attention.
Regarding cross-attention, it includes from tokens (as quries) to the image embedding and from image embedding to tokens (as keys and values).
After that, image embedding is upscaled by two transposed conv layers, and the foreground mask token is selected to perform point-wise product with the upscaled embedding to get the mask.
In comparison, AutoSAM deletes the prompt token in auxiliary embeddings so that it is not a promptable model any more.
The other modification is the duplication of auxiliary embeddings and image embedding by the number of classes to generate masks for multiple classes.
The computation of each pair can be conducted in parallel thus the overhead associated with generating extra mask is ignorable.
An alternative way to generate multiple masks for one inference is simply adding more foreground mask tokens in the output tokens.
However, we choose the first strategy because, intuitively, one group of auxiliary embeddings represents one object to be segmented in SAM. 
AutoSAM enables generation masks for each class independently.

\textbf{Convolutional Neural Network}
This type of prediction head is the representation of decoder in many popular segmentation model for medical images, like UNet\cite{ronneberger2015u}, UNet++\cite{zhou2019unet++}, TransUNet\cite{chen2021transunet}, and Swin-UNetr\cite{hatamizadeh2021swin}.
We firstly reshape the image embedding to a feature map of size (256, 64, 64).
Following the structure in UNet, the CNN head has $k$ stages ($k >= 2$), and each stage consists of conv layer with stride being 1 and transposed conv layer with stride being 2 to upscale.
 Different value of $k$ are tried in the experiment part, and when $k > 2$, the transposed conv layer is replaced with conv layer in $k-2$ stage(s), so that the output feature maps are always upscaled by 4x.
 Finally, a point-wise conv layer with kernel size 1 is applied to produce prediction masks for each class.

\textbf{Linear Layer}
The simple classification head is always used to evaluate the generalization of feature representation learned in the pre-training task \cite{chen2020simple, he2020momentum,he2022masked, oquab2023dinov2}.
In this work, we also apply a linear head to test if there is high-level semantic information extracted by the SAM encoder.
Same as CNN head, we resahpe the image embedding to a 2D feature map, and then directly deploy two transpose conv layers.
After that, we use two conv layers with kernel size 1 in replace of MLP to get the classification for each pixel.

\newcolumntype{Y}{>{\centering\arraybackslash}X}
\begin{table*}[t]
    \small
    \centering
    \caption{Comparisons of different methods trained with different number of labeled volumes on ACDC dataset. The three classes to be segmented are, "RV" (right ventricle), "Myo" (myocardium), and "LV"(left ventricle)."unsup" means no fine-tuning stage and the mask is generated based on given box prompts.}
    \label{table1}
    \begin{tabularx}{\linewidth}{c|c|Y|Y|Y|Y|Y}
    \toprule \hline
 \multicolumn{1}{c|}{} &\multirow{2}{*}{Methods} &\multicolumn{4}{c|}{Dice\% $\uparrow$} &\multirow{2}{*}{ASSD $\downarrow$} \\ \cline{3-6}
 & &RV &Myo &LV &avg  \\
 \hline

 \multirow{6}{*}{sup w/} &UNET   &13.45$\pm$1.89  &16.24$\pm$4.14  &22.95$\pm$0.47   &17.55$\pm$2.05 & 51.55$\pm$6.42  \\
 \multirow{6}{*}{\textbf{1} volume} &UNET + SimCLR   &14.25$\pm$6.52  &19.40$\pm$6.36  &27.54$\pm$9.80   &20.40$\pm$3.95 & 33.14$\pm$4.39  \\
  &Encoder + LN  &0.00$\pm$0.00  &20.42$\pm$13.20  &48.40$\pm$22.50   &22.94$\pm$12.32 &49.38$\pm$12.32   \\
 &Encoder + CNN   &30.66$\pm$14.28  &\textbf{39.96$\pm$8.14}  &50.55$\pm$13.56    &\textbf{40.39$\pm$11.90} &38.13$\pm$16.42 \\
 &AutoSAM (ft all)   &17.10$\pm$9.76  &30.05$\pm$7.77 &43.82$\pm$13.91    &30.32$\pm$10.05 &25.93$\pm$1.94 \\
  &AutoSAM   &\textbf{31.66$\pm$13.26}  &33.49$\pm$9.23  &\textbf{52.83$\pm$16.49}    &39.32$\pm$12.82 &\textbf{23.59$\pm$2.07} \\

 \hline \hline
 \multirow{6}{*}{sup w/} &UNET   &40.36$\pm$2.36  &52.23$\pm$3.80  &62.91$\pm$5.58   &51.83$\pm$3.41 &32.28$\pm$1.40  \\
 \multirow{6}{*}{\textbf{5} volumes} &UNET + SimCLR   &45.48$\pm$4.65  &58.20$\pm$6.12   &68.95$\pm$3.88 &57.18$\pm$3.20   &28.98$\pm$7.13  \\
 &Encoder + LN  &22.07$\pm$11.2  &37.38$\pm$11.56  &33.69$\pm$27.63   &31.05$\pm$16.14 &-  \\
 &Encoder + CNN   &\textbf{59.87$\pm$1.86}  &\textbf{62.81$\pm$2.82}  &78.96$\pm$2.79    &\textbf{67.21$\pm$1.32} &25.46$\pm$11.14 \\
 &AutoSAM (ft all)   &22.43$\pm$18.03  &37.08$\pm$13.49  &53.75$\pm$15.08    &37.76$\pm$15.22  &24.44$\pm$9.92 \\
  &AutoSAM   &58.48$\pm$3.90  &62.18$\pm$2.97  &\textbf{80.58$\pm$1.42}    &67.08$\pm$2.56 &\textbf{17.54$\pm$3.65} \\
  \hline \hline

 \multirow{1}{*}{unsup} &SAM (box)    &53.57$\pm$0.86 &39.60$\pm$0.65 &0.00$\pm$0.00   &31.06$\pm$0.41 &7.83$\pm$0.67   \\ \hline

\bottomrule
\end{tabularx}
\end{table*}

\section{Experiment}
\label{sec:exp}

\subsection{Dataset}
\textbf{ACDC Dataset} 
The ACDC (Automated Cardiac Diagnosis Challenge)\cite{bernard2018deep} dataset is part of the MICCAI 2017 challenges, which contains MRI scans of cardiac structures from 100 patients, each with 2 3D volumes. This dataset also provides expert segmentation masks of the left ventricle, right ventricle, and myocardium. 
We randomly split the MRI scans on patient basis into three parts,training set, validation set, and test set, with ratio being 70:15:15.

For preprocessing, we do normalization for each volume so that all pixels in a volume are zero mean and unit variance.
We then convert the pixel value to RGB format, and store each slice within the volume as PNG files, since SAM are trained on RGB images and we aim to keep the input format consistent.
Until then, although the MRI scans are given in 3D volumes, the segmentation is conducted on 2D images 
We compute the dice score as well as average symmetric surface distance (ASSD) for each volume in the test set, then regenerate the splits and repeat the experiments. 
The average score and the standard deviation for the four runs are reported.

\subsection{Training Recipe}

The implementation of training is based on deep learning package PyTorch. 
The GPU device used is a NVIDIA Tesla V100 with 16GB memory, which is more accessible than A100.
In comparison, SAM distribute training across 256 A100 GPUs.
During training, we randomly apply data augmentations on the input images including, Gaussian noise, brightness modification, elastic deformation, and rotation.
The training loss is a combination of Cross-Entropy loss and dice loss.
The optimizer algorithm used for updating is based on the Adam\cite{kingma2014adam}.
The learning rate is set as 0.0005 with $(\beta_{1}, \beta_{2}) = (0.5, 0.999)$.
The maximum batch size for one single GPU is 4 for all three prediction heads.
The default training epoch is 120 because we observe convergence of losses on validation set after that number of epochs.

\subsection{Baselines}
To validate the effectiveness of our proposed method, we
conduct experiments with some baseline approaches under the
same setting as comparison.
The first one is training a UNet from scratch, the most common method to get a automatic segmentation model for a specific dataset.
Secondly, We also try a self-supervised learning method, SimCLR\cite{chen2020simple}, which is widely used for label-efficient segmentation in medical image domain \cite{chaitanya2020contrastive, hu2021semi,zeng2021positional}.
This SimCLR baseline consists of two stages, pre-training and finetuning.
In the pr-training stage, we use all data in the training set without any annotation information.
We get two random views from the input images, and project them into feature space with encoder of UNet.
Then a contrastive loss is applied to maximize the agreement between the embeddings of the two views.
During the finetuning, the encoder of UNet is initialized with the pre-trained weights and all parameters in the model are trained on labeled data. 
Lastly, we try the original SAM without any finetuning to address the necessity of customizing SAM to a specific dataset.
Regarding the prompt, we use the box-style prompt, and the box coordinates are calculated based on the ground truth masks.

\begin{figure*}
    \centering
    \includegraphics[width=0.8\linewidth]{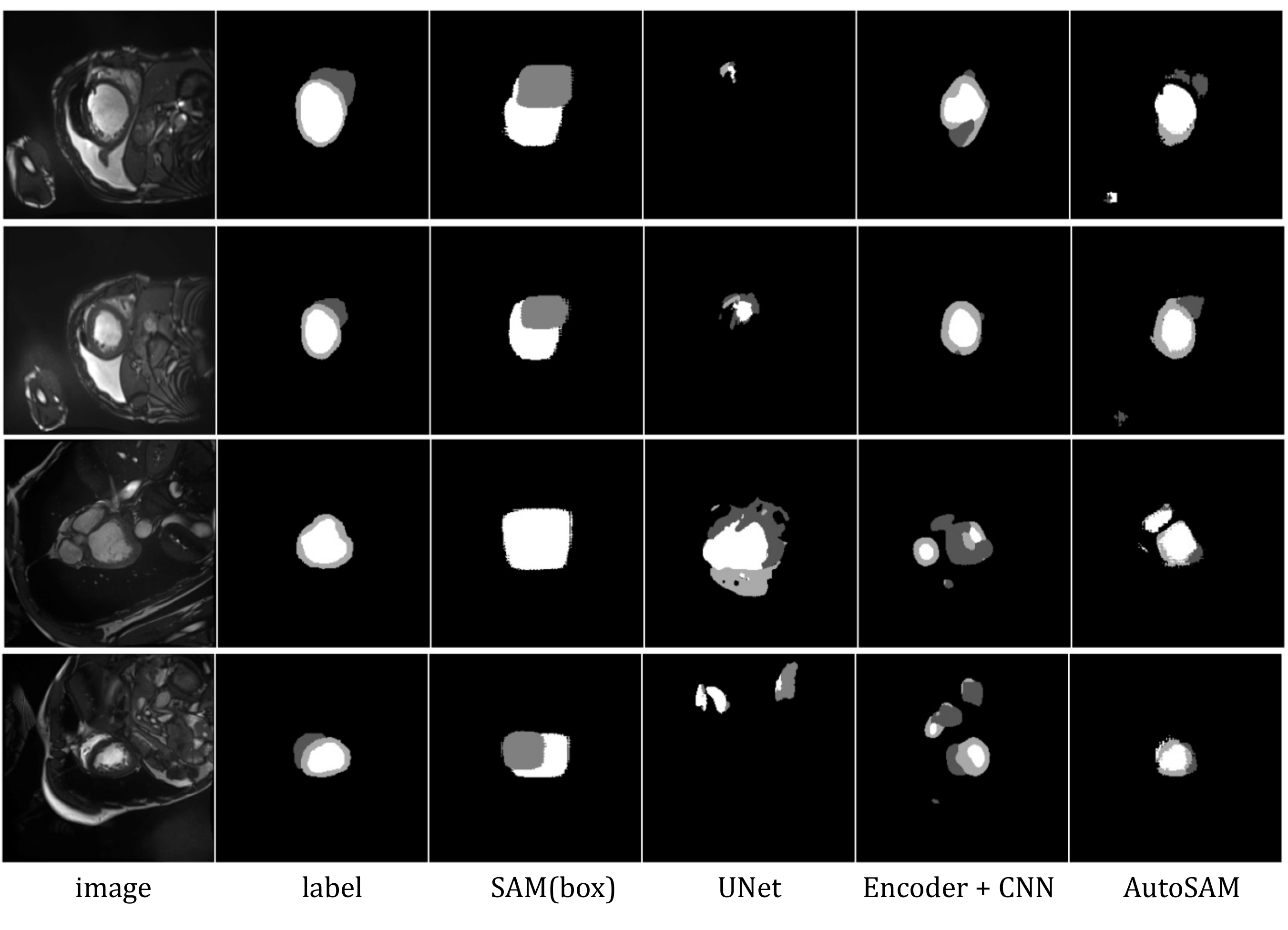}
    \caption{Visualization of prediction masks on ACDC dataset using different methods. SAM(box) is a zero-shot approach with only box-style prompts, and the prompts for three different classes are given at the same time. ``UNet", ``Encoder + CNN", and ``AutoSAM" are trained with only one labeld volume.}
    \label{fig4}
\end{figure*}

\subsection{Results}
\label{sec:abd_discussion}
\subsubsection{Label-efficient Adaptation} When finetuning a model on a new dataset,  to reduce the cost of annotating, it is desired that the finetuning achieves promising results with only limited annotated images.
Therefore, in Table \ref{table1}, we only provide 1 or 5 labeled volumes to evaluate the data-efficiency of our methods.
Here are the key observations drawn from Table \ref{table1}.

\textbf{i)} Firstly, AutoSAM and CNN head shows the best segmentation accuracy compared with all other methods for both settings. 
Especially when only prvoided with 1 labeled volume, the average dice score of AutoSAM is 39.32, almost twice higher than that of UNET and SimCLR.
This provides compelling evidence that the features learned in SAM encoder is general enough to be transferred to medical images.
In terms of statistical significance, it is hard to tell whether AutoSAM or CNN has higher dice score, why also implies that the strong power of SAM is mainly a consequence of representative features extracted by image encoder not the mask decoder.
Additionally, we observe that AutoSAM has lower ASSD compared to CNN head.
This difference can potentially be credited to the training of SAM decoder, which is designed to generate mask of a object concentrated near the prompt's location. 
In comparison, CNN head has no information loaded from SAM decoder, leading to higher ASSD values.

\textbf{ii)} Secondly, SAM shows worse segmentation performance compared with AutoSAM and CNN encoder even trained with only 1 volume, which strongly supports that finetuning SAM is an efficient way to solve its performance drop on a medical image dataset.
However, it is also noticed that, SAM has much lower ASSD than other methods.
This observation can be contributed to that SAM benefits from the provided localization information embedded in the box prompt. 
This localization information forces the prediction mask to be around box area.
On the other hand, the dice score of LV is always 0 for SAM.
 According to Fig.\ref{fig4}, we can find that Myo is a thin circle surrounded by the other two classes and the boundary is also blurred. 
 Since the box of Myo is close to the box of RB, the Myo is actually mistaken as part of RV, and consequently all LV area is predicted as Myo. 
 
 \begin{table}[t]
    \tiny
    \centering
    \caption{Ablation study of number of depths in the CNN prediction head given different sizes of labeled data. The reported metric is the average dice score on three classes.}
    \label{table2}
    \resizebox{\columnwidth}{!}{
    \begin{tabular}{ccccc}
    \toprule
 & 2 & 3 &4 & 5 \\
\hline
1 volume &31.46 &34.77 &40.39 &37.55 \\
5 volumes &59.56
 &62.06 &67.21 &65.48 \\
\bottomrule
\end{tabular}
}
\end{table}

\textbf{iii)} As is showed in Table \ref{table1}, the linear prediction head has substantially worse performance than the other two prediction heads.
Especially, when the number of labeled data is increased from 1 to 5, the linear head fails to gain much segmentation accuracy improvement.
We believe this outcome is due to the extreme light weight architecture.
When the visual features produced by SAM encoder do not have rich semantic information for medical images, such a simple prediction head results in weak model capabilities and may suffer from underfitting.

\subsubsection{Ablation Study}
The first ablation study we conduct is about how the number of depths in the CNN prediction head influences the finetuning results. 
In Table .\ref{table2}, the dice increases as the number of depths increases until depth=4. 
As is discussed above, a linear prediction head might suffer from underfitting,
When the depth is less than 4, larger prediction head leads to better model capabilities 
Nevertheless, when the number of depth exceeds 4, the benefits gained from increasing parameters in prediction head begin to diminish.
At this point, the quality of image embedding or prediction head architecture become more crucial factors in determining the performance.

We also evaluate the performance of AutoSAM and Encoder + CNN with different encoder sizes provided by SAM, which are vit-b, vit-l, and vit-h. 
Table. \ref{table3} shows that generally large model size leads to better finetuing results on the downstream tasks, but AutoSAM is less sensitive to encoder architecture than Encoder + CNN.
When using vit-h backbone, CNN head has significantly higher dice score than AutoSAM, though it still has higher ASSD.
Table. \ref{table3} can also serve as a reference regarding trad-off between efficiency and performance, as vit-h results in longer finetuning time and higher inference latency compared with vit-b.

\begin{figure}
    \centering
    \includegraphics[width=\linewidth]{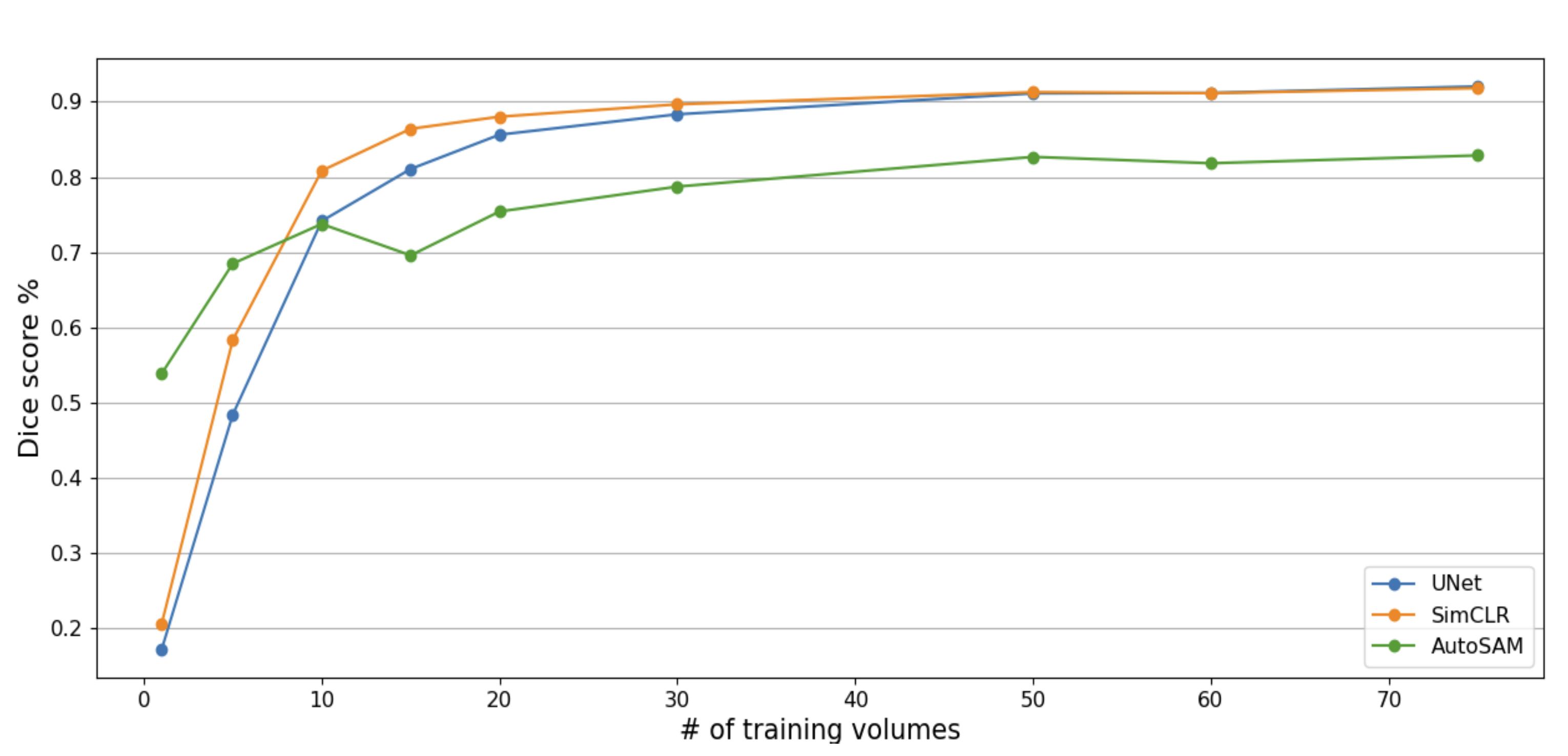}
    \caption{The change of dice score with respect to the number of labeled data in training. The results of UNet, UNet + SimCLR, and AutoSAM are included. Best viewed in colors.}
    \label{fig5}
\end{figure}

Lastly, we plot the results of using more labeled data for finetuning in Fig. \ref{fig5}. 
We find that AutoSAM only has advantages over UNet (with no additional information) and SimCLR (pretrained knowledge on the same dataset) when the labeled volume number is less than 10.
This is because SAM is pretrained on a large scale image dataset and the image encoder is capable of extracting semantic information, which can  benefit downstream segmentation task.
However, since SAM has never been exposed to medical images, this semantic information can be biased and specific to natural images.
It seems that with enough annotated data, the knowledge obtained from natural images poses negative impact when adapting prediction head specifically to medical image domain.
Therefore, to establish a real ``foundation model" for all image modalities, a large scale medical image dataset is in need for pretraining SAM in the future.

\begin{table}[t]
    \tiny
    \centering
    \caption{Influence of different image encoder size on the finetuning results. The three model type, vit-b, vit-l, and vit-h stand for the size of vision trasnformer encoder. Vit-b is the smallest model, and vit-h is the largest model. This ablation study is conducted on the setting with 1 labeled data for supervision. The dice score and assd are averaged across different folds. }
    \label{table3}
    \resizebox{0.9\columnwidth}{!}{
    \begin{tabular}{c|c|c|c}
    \toprule
 Model &Method & Dice\% & ASSD \\
\hline
UNet &SimCLR &20.40 &33.14 \\
\hline
vit-b &\multirow{3}{*}{AutoSAM} & 39.32 & 23.59 \\
vit-l & &40.08 &\textbf{18.00}  \\
vit-h & & \textbf{41.58} & 18.28 \\
\hline
vit-b &\multirow{3}{*}{Encoder + CNN} & 40.39 & 38.13 \\
vit-l & &42.80 &33.64 \\
vit-h & & \textbf{45.42} & \textbf{31.57} \\

\bottomrule
\end{tabular}
}
\end{table}

\section{Discussion}
\subsection{Conclusion}
Despite the success of SAM in natural images, how to efficiently adapt SAM to out-of-distribution medical image datasets remains an open-ended question.
Different from existing works, this paper provides a new perspective to solve this problem, that is freezing the weights in SAM image encoder and appending a light-weight task-specific prediction head.
To promote widespread applications, we modify SAM to be non-promptable and be able to generate multi-class masks.
We explore three types of prediction heads, ViT (called AutoSAM), CNN, and linear layer, in which AutoSAM and CNN head shows promising results in a few shot learning setting.
The fact that finetuned with only one labeled volume has better peformance than box-prompted SAM demonstrates the necessity of customizing SAM for a new dataset.
With limited number of annotated volumes, our methods is superior to training from scratch and self-supervised learning baselines.

\subsection{Future works}
Please note that this project is still ongoing, and there are several future directions we plan to explore.
Firstly, we intend to evaluate our method on more medical image datasets to verify the generalization of our findings. 
We will try medical image dataset with different modalities and different objects.
Furthermore, we recognize that there is still space for improvement for finetuning SAM using a limited number of volumes, such as one or five, compared to utilizing all available labeled data for training.
We will try prediction heads with more complex architecture, like DeepLabV3\cite{deeplabv3plus2018}, PSANet\cite{zhao2018psanet}, and UPerNET\cite{xiao2018unified}.
Lastly, we aim to include more baselines in our evaluations, which will involve comparing with other segmentation models and different self-supervised learning approaches.

{\small
\bibliographystyle{ieee_fullname}
\bibliography{11_references}

\begin{thebibliography}{10}\itemsep=-1pt

\bibitem{multi-atlas}
“multi-atlas labeling beyond the cranial vault.
\newblock \url{https://www.synapse.org/\#!Synapse:syn3193805/wiki/89480}.
\newblock Accessed: 2021-09-06.

\bibitem{bernard2018deep}
Olivier Bernard, Alain Lalande, Clement Zotti, Frederick Cervenansky, Xin Yang,
  Pheng-Ann Heng, Irem Cetin, Karim Lekadir, Oscar Camara, Miguel
  Angel~Gonzalez Ballester, et~al.
\newblock Deep learning techniques for automatic mri cardiac multi-structures
  segmentation and diagnosis: is the problem solved?
\newblock {\em IEEE transactions on medical imaging}, 37(11):2514--2525, 2018.

\bibitem{brown2020language}
Tom Brown, Benjamin Mann, Nick Ryder, Melanie Subbiah, Jared~D Kaplan, Prafulla
  Dhariwal, Arvind Neelakantan, Pranav Shyam, Girish Sastry, Amanda Askell,
  et~al.
\newblock Language models are few-shot learners.
\newblock {\em Advances in neural information processing systems},
  33:1877--1901, 2020.

\bibitem{chaitanya2020contrastive}
Krishna Chaitanya, Ertunc Erdil, Neerav Karani, and Ender Konukoglu.
\newblock Contrastive learning of global and local features for medical image
  segmentation with limited annotations.
\newblock {\em Advances in Neural Information Processing Systems},
  33:12546--12558, 2020.

\bibitem{chen2021transunet}
Jieneng Chen, Yongyi Lu, Qihang Yu, Xiangde Luo, Ehsan Adeli, Yan Wang, Le Lu,
  Alan~L Yuille, and Yuyin Zhou.
\newblock Transunet: Transformers make strong encoders for medical image
  segmentation.
\newblock {\em arXiv preprint arXiv:2102.04306}, 2021.

\bibitem{deeplabv3plus2018}
Liang-Chieh Chen, Yukun Zhu, George Papandreou, Florian Schroff, and Hartwig
  Adam.
\newblock Encoder-decoder with atrous separable convolution for semantic image
  segmentation.
\newblock In {\em ECCV}, 2018.

\bibitem{chen2020simple}
Ting Chen, Simon Kornblith, Mohammad Norouzi, and Geoffrey Hinton.
\newblock A simple framework for contrastive learning of visual
  representations.
\newblock In {\em International conference on machine learning}, pages
  1597--1607. PMLR, 2020.

\bibitem{dosovitskiy2020image}
Alexey Dosovitskiy, Lucas Beyer, Alexander Kolesnikov, Dirk Weissenborn,
  Xiaohua Zhai, Thomas Unterthiner, Mostafa Dehghani, Matthias Minderer, Georg
  Heigold, Sylvain Gelly, et~al.
\newblock An image is worth 16x16 words: Transformers for image recognition at
  scale.
\newblock {\em arXiv preprint arXiv:2010.11929}, 2020.

\bibitem{hatamizadeh2021swin}
Ali Hatamizadeh, Vishwesh Nath, Yucheng Tang, Dong Yang, Holger~R Roth, and
  Daguang Xu.
\newblock Swin unetr: Swin transformers for semantic segmentation of brain
  tumors in mri images.
\newblock In {\em International MICCAI Brainlesion Workshop}, pages 272--284.
  Springer, 2021.

\bibitem{he2022masked}
Kaiming He, Xinlei Chen, Saining Xie, Yanghao Li, Piotr Doll{\'a}r, and Ross
  Girshick.
\newblock Masked autoencoders are scalable vision learners.
\newblock In {\em Proceedings of the IEEE/CVF Conference on Computer Vision and
  Pattern Recognition}, pages 16000--16009, 2022.

\bibitem{he2020momentum}
Kaiming He, Haoqi Fan, Yuxin Wu, Saining Xie, and Ross Girshick.
\newblock Momentum contrast for unsupervised visual representation learning.
\newblock In {\em Proceedings of the IEEE/CVF conference on computer vision and
  pattern recognition}, pages 9729--9738, 2020.

\bibitem{he2023accuracy}
Sheng He, Rina Bao, Jingpeng Li, P~Ellen Grant, and Yangming Ou.
\newblock Accuracy of segment-anything model (sam) in medical image
  segmentation tasks.
\newblock {\em arXiv preprint arXiv:2304.09324}, 2023.

\bibitem{hu2021semi}
Xinrong Hu, Dewen Zeng, Xiaowei Xu, and Yiyu Shi.
\newblock Semi-supervised contrastive learning for label-efficient medical
  image segmentation.
\newblock In {\em Medical Image Computing and Computer Assisted
  Intervention--MICCAI 2021: 24th International Conference, Strasbourg, France,
  September 27--October 1, 2021, Proceedings, Part II 24}, pages 481--490.
  Springer, 2021.

\bibitem{jia2021scaling}
Chao Jia, Yinfei Yang, Ye Xia, Yi-Ting Chen, Zarana Parekh, Hieu Pham, Quoc Le,
  Yun-Hsuan Sung, Zhen Li, and Tom Duerig.
\newblock Scaling up visual and vision-language representation learning with
  noisy text supervision.
\newblock In {\em International Conference on Machine Learning}, pages
  4904--4916. PMLR, 2021.

\bibitem{kingma2014adam}
Diederik~P Kingma and Jimmy Ba.
\newblock Adam: A method for stochastic optimization.
\newblock {\em arXiv preprint arXiv:1412.6980}, 2014.

\bibitem{kirillov2023segment}
Alexander Kirillov, Eric Mintun, Nikhila Ravi, Hanzi Mao, Chloe Rolland, Laura
  Gustafson, Tete Xiao, Spencer Whitehead, Alexander~C Berg, Wan-Yen Lo, et~al.
\newblock Segment anything.
\newblock {\em arXiv preprint arXiv:2304.02643}, 2023.

\bibitem{lin2014microsoft}
Tsung-Yi Lin, Michael Maire, Serge Belongie, James Hays, Pietro Perona, Deva
  Ramanan, Piotr Doll{\'a}r, and C~Lawrence Zitnick.
\newblock Microsoft coco: Common objects in context.
\newblock In {\em Computer Vision--ECCV 2014: 13th European Conference, Zurich,
  Switzerland, September 6-12, 2014, Proceedings, Part V 13}, pages 740--755.
  Springer, 2014.

\bibitem{ma2023segment}
Jun Ma and Bo Wang.
\newblock Segment anything in medical images.
\newblock {\em arXiv preprint arXiv:2304.12306}, 2023.

\bibitem{oquab2023dinov2}
Maxime Oquab, Timoth{\'e}e Darcet, Th{\'e}o Moutakanni, Huy Vo, Marc
  Szafraniec, Vasil Khalidov, Pierre Fernandez, Daniel Haziza, Francisco Massa,
  Alaaeldin El-Nouby, et~al.
\newblock Dinov2: Learning robust visual features without supervision.
\newblock {\em arXiv preprint arXiv:2304.07193}, 2023.

\bibitem{ouyang2022training}
Long Ouyang, Jeffrey Wu, Xu Jiang, Diogo Almeida, Carroll Wainwright, Pamela
  Mishkin, Chong Zhang, Sandhini Agarwal, Katarina Slama, Alex Ray, et~al.
\newblock Training language models to follow instructions with human feedback.
\newblock {\em Advances in Neural Information Processing Systems},
  35:27730--27744, 2022.

\bibitem{radford2021learning}
Alec Radford, Jong~Wook Kim, Chris Hallacy, Aditya Ramesh, Gabriel Goh,
  Sandhini Agarwal, Girish Sastry, Amanda Askell, Pamela Mishkin, Jack Clark,
  et~al.
\newblock Learning transferable visual models from natural language
  supervision.
\newblock In {\em International conference on machine learning}, pages
  8748--8763. PMLR, 2021.

\bibitem{ramesh2021zero}
Aditya Ramesh, Mikhail Pavlov, Gabriel Goh, Scott Gray, Chelsea Voss, Alec
  Radford, Mark Chen, and Ilya Sutskever.
\newblock Zero-shot text-to-image generation.
\newblock In {\em International Conference on Machine Learning}, pages
  8821--8831. PMLR, 2021.

\bibitem{ronneberger2015u}
Olaf Ronneberger, Philipp Fischer, and Thomas Brox.
\newblock U-net: Convolutional networks for biomedical image segmentation.
\newblock In {\em International Conference on Medical image computing and
  computer-assisted intervention}, pages 234--241. Springer, 2015.

\bibitem{solaiman2019release}
Irene Solaiman, Miles Brundage, Jack Clark, Amanda Askell, Ariel Herbert-Voss,
  Jeff Wu, Alec Radford, Gretchen Krueger, Jong~Wook Kim, Sarah Kreps, et~al.
\newblock Release strategies and the social impacts of language models.
\newblock {\em arXiv preprint arXiv:1908.09203}, 2019.

\bibitem{wu2023medical}
Junde Wu, Rao Fu, Huihui Fang, Yuanpei Liu, Zhaowei Wang, Yanwu Xu, Yueming
  Jin, and Tal Arbel.
\newblock Medical sam adapter: Adapting segment anything model for medical
  image segmentation.
\newblock {\em arXiv preprint arXiv:2304.12620}, 2023.

\bibitem{xiao2018unified}
Tete Xiao, Yingcheng Liu, Bolei Zhou, Yuning Jiang, and Jian Sun.
\newblock Unified perceptual parsing for scene understanding.
\newblock In {\em Proceedings of the European Conference on Computer Vision
  (ECCV)}, pages 418--434, 2018.

\bibitem{zeng2021positional}
Dewen Zeng, Yawen Wu, Xinrong Hu, Xiaowei Xu, Haiyun Yuan, Meiping Huang, Jian
  Zhuang, Jingtong Hu, and Yiyu Shi.
\newblock Positional contrastive learning for volumetric medical image
  segmentation.
\newblock In {\em Medical Image Computing and Computer Assisted
  Intervention--MICCAI 2021: 24th International Conference, Strasbourg, France,
  September 27--October 1, 2021, Proceedings, Part II 24}, pages 221--230.
  Springer, 2021.

\bibitem{zhang2023customized}
Kaidong Zhang and Dong Liu.
\newblock Customized segment anything model for medical image segmentation.
\newblock {\em arXiv preprint arXiv:2304.13785}, 2023.

\bibitem{zhao2018psanet}
Hengshuang Zhao, Yi Zhang, Shu Liu, Jianping Shi, Chen Change~Loy, Dahua Lin,
  and Jiaya Jia.
\newblock Psanet: Point-wise spatial attention network for scene parsing.
\newblock In {\em Proceedings of the European Conference on Computer Vision
  (ECCV)}, pages 267--283, 2018.

\bibitem{zhou2017scene}
Bolei Zhou, Hang Zhao, Xavier Puig, Sanja Fidler, Adela Barriuso, and Antonio
  Torralba.
\newblock Scene parsing through ade20k dataset.
\newblock In {\em Proceedings of the IEEE conference on computer vision and
  pattern recognition}, pages 633--641, 2017.

\bibitem{zhou2019unet++}
Zongwei Zhou, Md~Mahfuzur~Rahman Siddiquee, Nima Tajbakhsh, and Jianming Liang.
\newblock Unet++: Redesigning skip connections to exploit multiscale features
  in image segmentation.
\newblock {\em IEEE transactions on medical imaging}, 39(6):1856--1867, 2019.

\end{thebibliography}
}

\ifarxiv \clearpage \appendix
\label{sec:appendix}

 \fi

\end{document}


\title{\paperTitle \\ Supplemental Material}
\author{\authorBlock}
\maketitle

\appendix
\label{sec:appendix}


{\small
\bibliographystyle{ieee_fullname}
\bibliography{11_references}
}